\documentclass[letterpaper]{article} 
\usepackage{aaai2026}  
\usepackage{times}  
\usepackage{multirow}
\usepackage{adjustbox} 
\usepackage{array}
\usepackage{helvet}  
\usepackage{courier}  
\usepackage[hyphens]{url}  
\usepackage{graphicx} 
\usepackage{natbib}  
\usepackage{caption} 
\usepackage{algorithm}
\usepackage{algorithmic}

\usepackage{newfloat}
\usepackage{listings}

\usepackage{amsfonts}
\usepackage{amsmath}

\DeclareCaptionStyle{ruled}{labelfont=normalfont,labelsep=colon,strut=off} 
\floatstyle{ruled}
\newfloat{listing}{tb}{lst}{}
\floatname{listing}{Listing}
\title{Multi-Modal Assistance for Unsupervised Domain Adaptation
on Point Cloud 3D Object Detection}
\author{
    Shenao Zhao\equalcontrib \quad\quad
    Pengpeng Liang\equalcontrib\thanks{Corresponding author.}\quad\quad
    Zhoufan Yang
}
\affiliations{
    School of Computer Science and Artificial Intelligence, Zhengzhou University\\
    1722236579@qq.com\quad liangpcs@gmail.com \quad 2829149697@qq.com
}

\usepackage{bibentry}

\begin{document}

\maketitle

\begin{abstract}
Unsupervised domain adaptation for LiDAR-based 3D object detection (3D UDA) based on the teacher-student architecture with pseudo labels has achieved notable improvements in recent years. Although it is quite popular to collect point clouds and images simultaneously, little attention has been paid to the usefulness of image data in 3D UDA  when training the models. In this paper, we propose an approach named MMAssist that improves the performance of 3D UDA with multi-modal assistance. A method is designed to align 3D features between the source domain and the target domain by using image and text features as bridges. More specifically, we project the ground truth labels or pseudo labels to the images to get a set of 2D bounding boxes. For each 2D box, we extract its image feature from a pre-trained vision backbone. A large vision-language model (LVLM) is adopted to extract the box's text description, and a pre-trained text encoder is used to obtain its text feature. During the training of the model in the source domain and the student model in the target domain, we align the 3D features of the predicted boxes with their corresponding image and text features, and the 3D features and the aligned features are fused with learned weights for the final prediction. The features between the student branch and the teacher branch in the target domain are aligned as well. To enhance the pseudo labels, we use an off-the-shelf 2D object detector to generate 2D bounding boxes from images and estimate their corresponding 3D boxes with the aid of point cloud, and these 3D boxes are combined with the pseudo labels generated by the teacher model. Experimental results show that our approach achieves promising performance compared with state-of-the-art methods in three domain adaptation tasks on three popular 3D object detection datasets. The code is available at \url{https://github.com/liangp/MMAssist}.          
\end{abstract}
    
\section{Introduction}
\label{sec:intro}
3D object detection from point clouds is an important task in autonomous driving, and it has made significant progress in recent years with the availability of large-scale annotated data~\cite{caesar2020nuscenes,sun2020scalability}. However, due to the discrepancies between domains caused by varying LiDAR beams, diverse environments, etc, 3D object detectors trained in the source domain often encounter difficulties when applied to the target domain~\cite{chang2024cmda,wang2020train,wei2022lidar,zhang2024pseudo}.

\begin{figure}[t]
    \centering
    \includegraphics[width=\columnwidth]{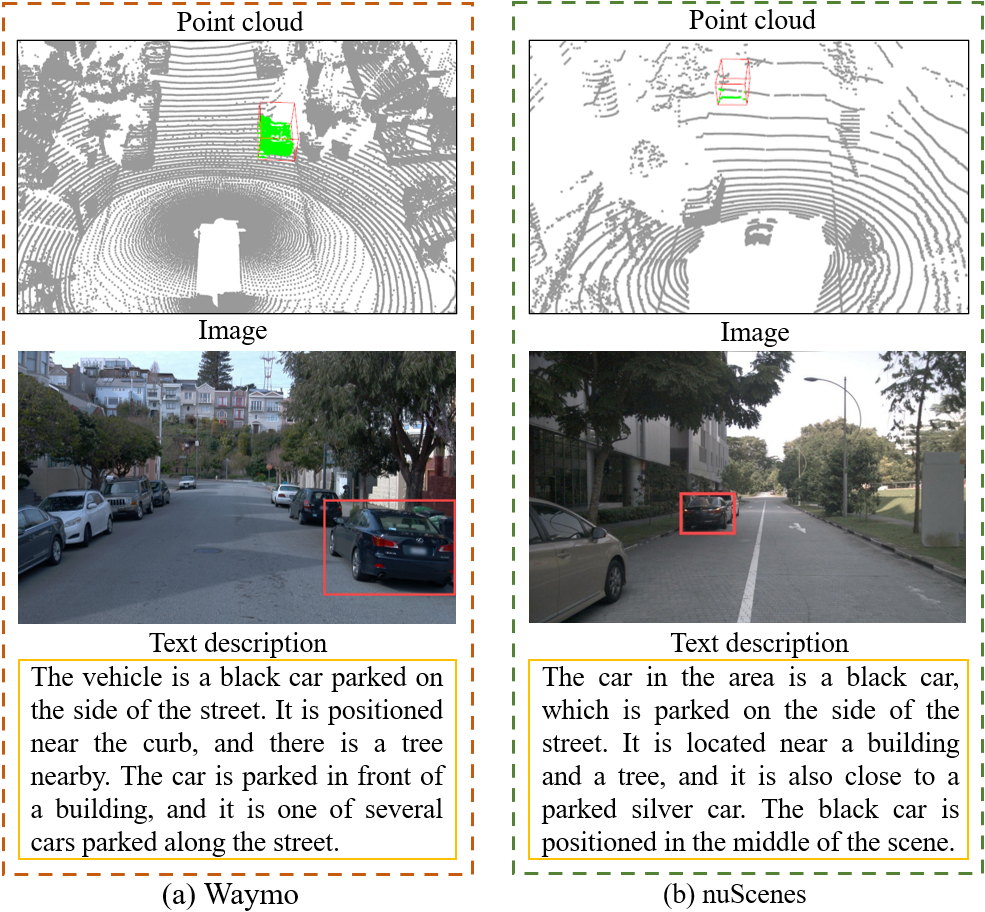}
    \vspace{-16pt}
    \caption{Comparison of point cloud, image, and object's text description generated by LLaVA~\cite{liu2023visual} of two objects having similar appearances in the images. (a) is from Waymo~\cite{sun2020scalability}, and (b) is from nuScenes~\cite{caesar2020nuscenes}. Best viewed in color.}
    \vspace{-20pt}
    \label{fig:soure_target}
\end{figure}

Unsupervised domain adaptation aims at transferring the model trained in the source domain to the target domain, in which no data is annotated. To improve the performance of LiDAR-based 3D UDA, one strategy is to reduce the data or feature distribution gap between the two domains~\cite{wang2020train,hu2023density}.  SN~\cite{wang2020train} normalizes the source domain data based on the difference in mean car sizes between the two domains. DTS~\cite{hu2023density} uses random beam re-sampling to augment the data of both domains to reduce the beam density gap.  Another strategy is to enhance the pseudo label quality under the self-training framework, which uses the source domain model or a teacher model to generate pseudo labels for the target domain~\cite{yang2022st3d++,zhang2024pseudo}. ST3D++~\cite{yang2022st3d++} designs a hybrid quality-aware triplet memory to improve the quality of pseudo labels. PERE~\cite{zhang2024pseudo} mitigates the impact of unreliable pseudo labels with box replacement and point removal. 


It is a common practice to collect LiDAR point cloud and image data simultaneously in autonomous driving~\cite{caesar2020nuscenes,sun2020scalability,geiger2012we}, but most of the current state-of-the-art approaches~\cite{hu2023density,shin2024towards,yihan2021learning,zhang2024pseudo,yang2022st3d++} solely depend on the LiDAR data without exploring the assistance of image data. CMDA~\cite{chang2024cmda} transfers semantic knowledge from images to the 3D detector of the source domain, but the usefulness of images is not directly exploited in the self-training of the target domain model.   Besides image data, the recent progress of large vision-language model (LVLM)~\cite{liu2023visual,wang2024qwen2} enables the extraction of accurate text descriptions of objects from images. Although LVLM has been successfully adopted by point cloud pre-training~\cite{xue2024ulip}, its effectiveness for LiDAR-based 3D UDA has not been well studied.  

In this paper, we present MMAssist that assists LiDAR-based 3D UDA with multi-modal information (image and text), and our approach is based on the teacher-student self-training method DTS~\cite{hu2023density}.  As vision models trained with large-scale data show strong generalization ability~\cite{lavoie2025large}, compared with the large discrepancy with regard to point clouds, image features of similar objects in different domains are expected to have a smaller domain gap. Meanwhile,  LVLM generates similar text descriptions for similar objects in two domains as shown in Fig.~\ref{fig:soure_target}.  So, instead of directly aligning features between two domains, which is difficult when the training processes of two domains are separate, we propose to use image and text features as bridges to align features across two domains. Given the ground truth labels in the source domain and pseudo labels in the target domain, we project the 3D bounding boxes to images to generate a set of 2D bounding boxes. For each 2D bounding box, we use RoIAlign~\cite{he2017mask} to extract its image feature from a pre-trained vision backbone and use an LVLM to generate its text description, then the text description is encoded into the text feature with a pre-trained text encoder. During the training in the source domain and the training of the student model in the target domain, when a predicted 3D bounding box matches a ground truth label (or pseudo label), we extract its 3D feature from the detector backbone and align it with its corresponding image and text features by minimizing losses based on cosine similarity. We fuse the 3D feature and its aligned features in a weighted manner to predict the final 3D bounding box. In the target domain, when a detected 3D box of the student branch matches a detected 3D box of the teacher branch, their 3D features are aligned as well.  

To improve the quality of pseudo labels, following~\cite{zhang2024approaching}, we use the off-the-shelf 2D object detector GroundingDINO~\cite{liu2024grounding} to generate 2D bounding boxes from images and lift them to 3D space with geometric reasoning~\cite{wei2021fgr}. Based on the observation that the 3D object detector trained in the source domain, in general, generates more accurate pseudo labels than the pseudo labels reasoned from 2D bounding boxes, but it has difficulty in detecting long-range 3D objects, we add 3D pseudo labels from images in the long-range area where the distance is larger than a threshold. And when there are two pseudo labels generated by the above two approaches with overlap larger than a threshold, we keep the pseudo labels generated by the 3D detector while discarding the other.

We follow~\cite{hu2023density} and integrate our approach into three popular LiDAR-based object detectors (PV-RCNN~\cite{shi2020pv}, SECOND-IoU~\cite{yang2021st3d}, and PointPillars~\cite{lang2019pointpillars}), and their performances are evaluated on three domain adaptation tasks using nuScenes~\cite{caesar2020nuscenes}, KITTI~\cite{geiger2012we}, and Waymo~\cite{sun2020scalability}.  The combination of three detectors and the three domain adaptation tasks results in nine subtasks, and our approach performs best in seven of them in general.

The contributions of our paper are summarized as follows:
\begin{itemize}
    \item We propose to assist in LiDAR-based 3D UDA with multi-modal information with respect to both cross-domain feature alignment and pseudo label quality.
    \item We design a method that uses image and text features as bridges for feature alignment between the source domain and the target domain.
    \item The performance of MMAssist is inspiring, and the effectiveness of each component of our approach is validated via ablation study. 
\end{itemize}

\section{Related work}
\label{sec:related}

\subsection{Unsupervised Domain Adaptation for 3D Object Detection}
A popular unsupervised domain adaptation approach for point cloud-based 3D object detection is pseudo label-based self-training, and current research efforts are mainly focused on improving the quality of pseudo labels and aligning the data distributions or feature representations across domains. ST3D++~\cite{yang2022st3d++} applies random object scaling during the pre-training phase and designs a voting method based on historical pseudo labels to reduce pseudo label noise. To reduce the sensitivity regarding the beam density variation,  DTS~\cite{hu2023density} presents a random beam re-sampling module and aligns features between the teacher and student branches with an object graph alignment module.  PERE~\cite{zhang2024pseudo} refines pseudo labels by randomly removing or replacing them based on prediction confidences and aligns instance features across domains with a triplet loss. MLC-Net~\cite{luo2021unsupervised} trains the student branch with both source domain labels and target domain pseudo labels, and it enhances the consistency between the teacher and student branches at the point, instance, and batch normalization levels. ReDB~\cite{chen2023revisiting} generates reliable pseudo labels with cross-domain examination and downsampling based on overlapped box counting. CMDA~\cite{chang2024cmda} mixes the source point cloud and target point cloud during self-training and learns domain-invariant features via instance-level adversarial learning. DALI~\cite{lu2024dali} enhances the quality of pseudo labels generated by the source domain model via scaling the coordinates of target domain point clouds, and synthetic point clouds are used to improve the consistency between pseudo labels and their corresponding point clouds. 

In addition to the above self-training methods, \cite{wei2022lidar} generates low-beam pseudo point clouds through downsampling and transfers the model trained on the high-beam source domain to the low-beam target domain using knowledge distillation. 3D-CoCo~\cite{yihan2021learning} trains the target domain model with pseudo labels and updates the source domain model with ground-truth labels simultaneously, and cross-domain instance features are aligned through contrastive learning during the co-training process. CMT~\cite{chen2024cmt} constructs a hybrid domain that leverages both the source and target domains to align features.   GPA-3D~\cite{li2023gpa} aligns the BEV features of the source and target domains to the same set of geometry-aware prototypes to reduce domain discrepancy.  GroupEXP-DA~\cite{shin2024towards} clusters the labels into groups that are shared between domains and learns group-equivariant spatial features. GBlobs~\cite{malic2025gblobs} encodes point cloud neighborhoods with Gaussian blobs to improve the generalization ability of detectors.

\subsection{LiDAR-based 3D Object Detection}
LiDAR-based 3D object detection aims to localize and classify objects from sparse and unordered point clouds. Point-based methods~\cite{shi2019pointrcnn,yang20203dssd,shi2020point} directly operate on raw point clouds, and they typically employ hierarchical networks such as PointNet~\cite{qi2017pointnet} and PointNet++~\cite{qi2017pointnet++} to extract point-wise features. Voxel-based methods~\cite{zhou2018voxelnet,deng2021voxel,yan2018second,lang2019pointpillars,fan2022embracing,zhang2024safdnet} convert unstructured point clouds into regularized 2D/3D grids and carry out detection based on the voxel representation. Meanwhile, PV-RCNN~\cite{shi2020pv} leverages both point-based and voxel-based representations by combining keypoints with nearby voxels. 
\section{Method}
\subsection{Problem Statement and Overview}
\label{sec:overview}
\begin{figure*}[ht]
    \centering
    \includegraphics[width=\textwidth]{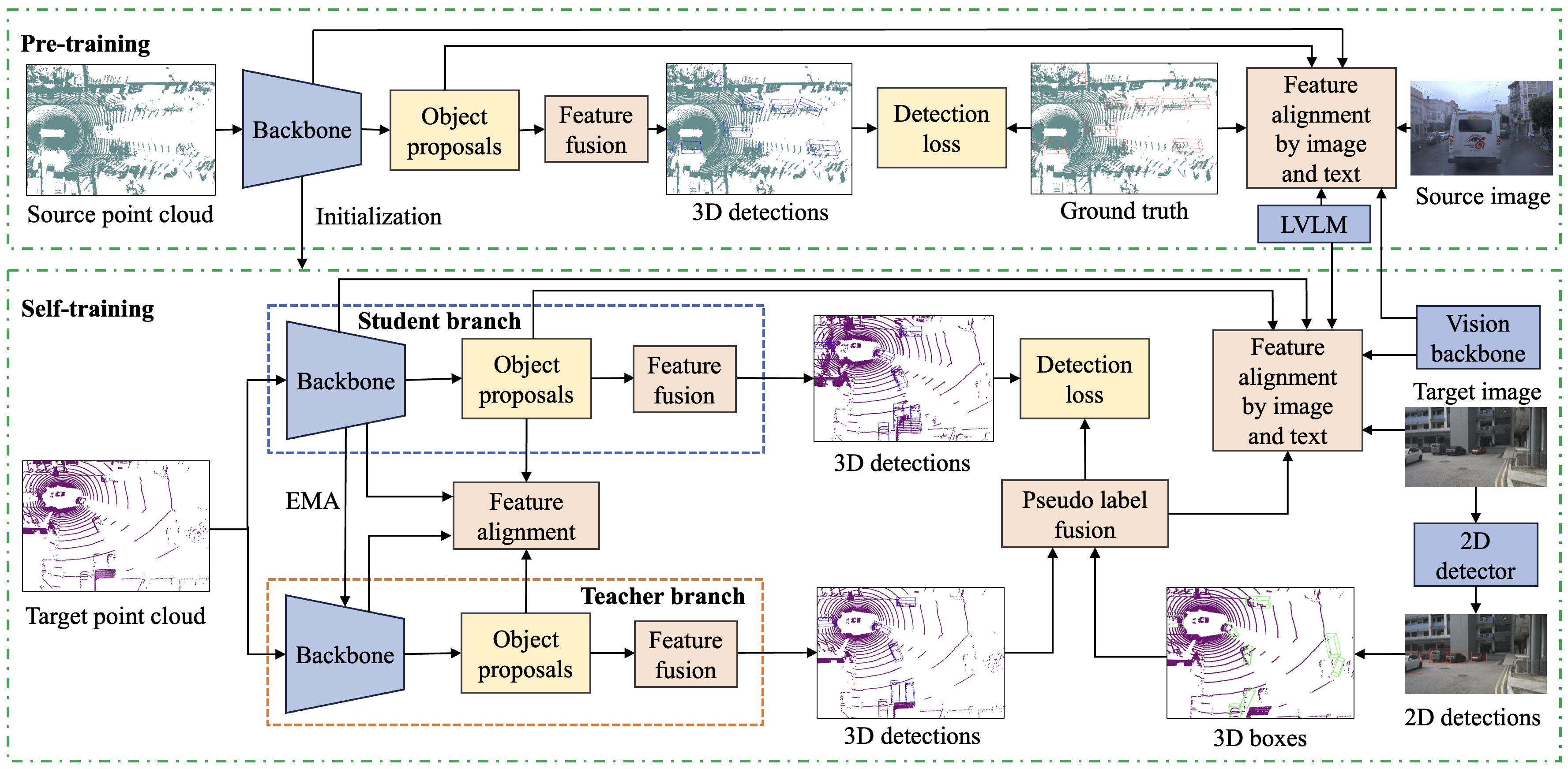}  
    \caption{Overall approach pipeline. Features of the source domain model and features of the student model in the target domain are aligned with the help of their corresponding image and text features. The aligned features are fused with the 3D features to make the final prediction (the input features to the feature fusion module are not shown). 3D bounding boxes estimated from the 2D detection results are used to enhance the pseudo labels. The teacher model is updated by the student model via EMA.}  
    \label{fig:framework}  
\vspace{-6pt}
\end{figure*}

\noindent\textbf{Problem statement.} Unsupervised domain adaptation for 3D object detection aims to adapt a model trained on a labeled source domain dataset 
$D^{S}=\{(P^S_i, Y^S_i)\}^{N_S}_{i=1}$ to an unlabeled target domain dataset $D^{T}=\{P^T_i\}^{N_T}_{i=1}$. $P^S_i$ and $Y^S_i$ represent the $i$-th point cloud sample from the source domain and its corresponding annotation, respectively. 
The label \(Y^S_i\)  includes object category information and 3D bounding box parameters. Each 3D bounding box is defined by its length $l$, width $w$, height $h$, center coordinate \((x, y, z)\), and heading angle \(\theta\). 
\(P^T_i\) denotes the \(i\)-th point cloud sample from the target domain, which does not have ground truth labels.

\noindent\textbf{Overview.} As shown in Fig. \ref{fig:framework}, our approach MMAssist  is based on the teacher-student framework. The source point cloud and the target point cloud of the student branch are augmented with beam re-sampling~\cite{hu2023density} so that they have similar densities. In the pre-training stage, we use $D^S$ to train the source model $M^{S}$. In the self-training stage, $M^{S}$ is used to initialize the teacher model $M^{\text{teacher}}$ and the student model $M^{\text{student}}$. We use pseudo labels to train $M^{\text{student}}$,  and the parameters of $M^{\text{teacher}}$ are updated with the parameters of $M^{\text{student}}$ using  exponential moving average (EMA), i.e., $M^{\text{teacher}}=\epsilon M^{\text{teacher}}+(1-\epsilon)M^{\text{student}}$. During both the pre-training stage and the self-training stage, we align the 3D features of objects with their corresponding image and text features to learn domain-invariant feature representations, which are further fused with the 3D features in a weighted manner to make the final prediction. In the self-training stage,  the pseudo labels generated by the teacher model and pseudo labels generated based on 2D detection results from images associated with $D^{T}$ are combined to train the student model. Features between the student branch and the teacher branch are aligned as well.  It is worth noting that image and text information is only used in the training stage, and the test stage solely uses point clouds as input.

\subsection{Cross-Domain Feature Alignment with Image and Text Features as Bridges}
\label{sec:Cross-DomainFeatureAlignmentwithTextasBridge}
In this section, we introduce the method for aligning features between two domains with image and text features as bridges. The alignment is performed in two domains individually. 

\noindent\textbf{Image and text features generation.}
Given a point cloud sample \( P \) from the source domain (or target domain) and its corresponding 3D ground truth labels (or pseudo labels) \( B = \{ \mathbf{b}_i = (x_i, y_i, z_i, l_i, w_i, h_i, \theta_i) \}_{i=1}^{N^{3D}} \), we use a camera's intrinsic matrix \( K \) and extrinsic matrix \( E \) to project each $\mathbf{b}_i$ to its 2D image plane.  We filter out the 2D projected bounding boxes outside the image. For the multi-camera setting, after projecting $\mathbf{b}_i$ to multiple image planes, we calculate the truncation rates of 2D projected bounding boxes and only retain the 2D box with the minimum truncation rate. We finally obtain a set of 2D bounding boxes \( B^{2D} = \{\mathbf{b}^{2D}_i = (x^{min}_i, y^{min}_i, x^{max}_i, y^{max}_i)\}_{i=1}^{N^{2D}} \), where $(x^{min}_i, y^{min}_i, x^{max}_i, y^{max}_i)$ indicates the coordinates of bounding box $\mathbf{b}_i^{2D}$'s top-left and bottom-right corners.

For each $\mathbf{b}_i$ in $B$ that has a corresponding $\mathbf{b}_{i}^{2D}$ in $B^{2D}$, we first use RoIAlign~\cite{he2017mask} with $\mathbf{b}_{i}^{2D}$ to extract its image feature $\mathbf{f}_{i}^{img}\in\mathbb{R}^{C^{img}}$ from the feature maps of $\mathbf{b}_{i}^{2D}$'s corresponding image $I$,  the feature maps of which are generated by a pre-trained vision backbone. Then,  we use a large vision-language model (LVLM) to extract its text description. More specifically, we construct a text prompt $p_i$=\textit{``There is a \{class\} in the area $(x^{min}_i, y^{min}_i, x^{max}_i, y^{max}_i)$, please describe the characteristics of this \{class\}."}, where \textit{class} is the category of $\mathbf{b}_i$. Then, $p_i$ and the image $I$ in which $\mathbf{b}_i^{2D}$ is located are fed into the LVLM to produce the text description $d_i$. Next, we use a pre-trained text encoder to extract the text feature $\mathbf{f}_{i}^{text}\in\mathbb{R}^{C^{text}}$ for $d_i$. We choose the backbone of GroundingDINO~\cite{liu2024grounding}, LLaVA~\cite{liu2023visual}, and SLIP~\cite{mu2022slip} as the vision backbone,  LVLM, and text encoder, respectively.

\noindent\textbf{Aligning 3D feature with image and text features.} 
For both the training of the source model in the pre-training stage and the training of the student model in the self-training stage, given a point cloud sample \( P \), we first obtain the initial predicted bounding boxes  $\hat{B} = \{ \mathbf{\hat{b}}_i = (x_i, y_i, z_i, l_i, w_i, h_i, \theta_i) \}_{i=1}^{L}$. For each $\mathbf{\hat{b}}_i$, we extract its 3D feature $\hat{\mathbf{f}}_i^{3D}\in\mathbb{R}^{C^{3D}}$ from the backbone of the detector (detail about the feature extraction is in the supplementary material). Then, $\hat{\mathbf{f}}_i^{3D}$ is converted to $\hat{\mathbf{f}}_i^{img}\in\mathbb{R}^{C^{img}}$ and $\hat{\mathbf{f}}_i^{text}\in\mathbb{R}^{C^{text}}$ by MLP, and they are used for feature alignment with image and text features, respectively. 

To find the corresponding image and text features of $\hat{\mathbf{b}}_i$, which we align  $\hat{\mathbf{f}}_i^{img}$ and  $\hat{\mathbf{f}}_i^{text}$ with, we match \(\hat{\mathbf{b}}_i\) to a 3D ground truth bounding box using the following formula:
\begin{equation}
\label{eq:findmaxiou}
\mathbf{b}_m =  \underset{\mathbf{b}_j \in B}{\arg\max} \ \text{IoU}(\hat{\mathbf{b}}_i, \mathbf{b}_j)
\end{equation}
\(\mathbf{b}_m\) is the 3D ground truth bounding box in $B$ that has the highest IoU with \(\hat{\mathbf{b}}_i\) (ground truth is the pseudo labels in the self-training stage). If \( \text{IoU}(\hat{\mathbf{b}}_i, \mathbf{b}_m) \geq \mu \), we consider $(\hat{\mathbf{b}}_i, \mathbf{b}_m)$ a matched pair. And if $\mathbf{b}_m$ has a matched 2D bounding box $\mathbf{b}_m^{2D}$, we use $\mathbf{b}_m^{2D}$'s image feature $\mathbf{f}_m^{img}$ and text feature  $\mathbf{f}_m^{text}$ as $\hat{\mathbf{b}_i}$'s image and text features for alignment, respectively, i.e.,  $\hat{\mathbf{g}}_{i}^{img}=\mathbf{f}_m^{img}$ and $\hat{\mathbf{g}}_{i}^{text}=\mathbf{f}_m^{text}$. Finally, we construct a  subset $\hat{B'}= \{ \mathbf{\hat{b}}_i = (x_i, y_i, z_i, l_i, w_i, h_i, \theta_i) \}_{i=1}^{L'}$ of $\hat{B}$, where $L'\leq L$, and each $\hat{\mathbf{b}}_i$ in $\hat{B'}$ has its corresponding image feature $\hat{\mathbf{g}}_i^{img}$ and text feature $\hat{\mathbf{g}}_i^{text}$.

To align with image features, we first randomly generate a set of background 2D bounding boxes $B^{2D}_{bg} = \{\mathbf{b}^{bg}_i = (x^{min}_i, y^{min}_i, x^{max}_i, y^{max}_i)\}_{i=1}^{N^{bg}}$ which don't intersect with any 2D box in $B^{2D}$, and the image feature $\mathbf{g}_i^{bg}$ of each background box is extracted with RoIAlign as well.  Then, for each $\hat{\mathbf{b}}_i$ in $\hat{B'}$, we pull its $\hat{\mathbf{f}}_i^{img}$ derived from its 3D feature $\hat{\mathbf{f}}_i^{3D}$ closer to its image feature $\hat{\mathbf{g}}_i^{img}$ and push it away from image features of background boxes. The corresponding loss is formulated as:

\begin{align}
    \label{eq:alignimage}
    \mathcal{L}_\text{align}^{\text{img}} = \frac{1}{L'}\sum_{i=1}^{L'} 
    \max\bigg(&\frac{1}{N^{bg}}\sum_{j=1}^{N^{bg}}\text{sim}(\hat{\mathbf{f}}^{img}_i, \mathbf{g}^\text{bg}_{j}) \notag\\
    &- \text{sim}(\hat{\mathbf{f}}^{img}_i, \hat{\mathbf{g}}^\text{img}_i) + \sigma, 0\bigg)
\end{align}
where $\text{sim}(\mathbf{a}, \mathbf{b}) = \frac{\mathbf{a} \cdot \mathbf{b}}{\|\mathbf{a}\| \|\mathbf{b}\|}$ is the cosine similarity, and $\sigma$ is the margin.

To align with text feature, for each $\hat{\mathbf{b}}_i$, we minimize the distance between $\hat{\mathbf{f}}_i^{text}$ and its text feature $\hat{\mathbf{g}}_i^{text}$, where  $\hat{\mathbf{f}}_i^{text}$ is derived from $\hat{\mathbf{b}}_i$'s 3D feature. Below is the loss based on cosine similarity:
\begin{equation}
    \label{eq:aligntext}
    \mathcal{L}_\text{align}^{\text{text}} = \frac{1}{L'}\sum_{i=1}^{L'} \left(1 - \text{sim}(\hat{\mathbf{f}}^{text}_i, \hat{\mathbf{g}}^\text{text}_i)\right)
\end{equation}

The final loss function of the pre-training stage is a combination of the detector's original loss $\mathcal{L}_\text{det}$, $\mathcal{L}_\text{align}^{\text{text}}$, and $\mathcal{L}_\text{align}^{\text{img}}$:
\begin{equation}
\label{eq:loss-pre-training}
    \mathcal{L}_{\text{pre}} = \mathcal{L}_\text{det} + \alpha \mathcal{L}_\text{align}^{\text{text}} + \beta\mathcal{L}_\text{align}^{\text{img}}
\end{equation}

\noindent\textbf{Fusing 3D features with aligned features.} 
Given a 3D bounding box $\hat{\mathbf{b}}$'s 3D feature $\mathbf{f}^{3D}\in\mathbb{R}^{C^{3D}}$, image-aligned feature $\mathbf{f}^{img}\in\mathbb{R}^{C^{img}}$, and text-aligned feature  $\mathbf{f}^{text}\in\mathbb{R}^{C^{text}}$, we fuse them in a weighted manner. More specifically, we first use two MLPs to change the dimensionality of $\mathbf{f}^{img}$ and $\mathbf{f}^{text}$ to $C^{3D}$, respectively, then we concatenate the three feature vectors and input it to another MLP to learn a weight vector $\mathbf{w}\in\mathbb{R}^{3}$. The final fused feature is obtained by:
\begin{equation}
\label{eq:fusion features}
    \mathbf{f}^{fused} = \mathbf{w}_0\mathbf{f}^{3D} + \mathbf{w}_1\mathbf{f}^{img} + \mathbf{w}_2\mathbf{f}^{text}
\end{equation}
After getting $\mathbf{f}^{fused}$, for PV-RCNN~\cite{shi2020pv}, we use it to refine the proposal from the first stage; for PointPillar~\cite{lang2019pointpillars}, we use it to refine the initial detection result; for SECOND-IoU~\cite{yang2021st3d}, we use it to predict the IoU score in the second stage.

\subsection{Aligning 3D Features between Student and Teacher Branches}
\label{sec:Aligning3DFeaturesbetweenStudentandTeacherBranches}
In the self-training stage, we follow~\cite{hu2023density} and further align the 3D features extracted from the student model and the teacher model. Given the predicted bounding boxes $\hat{B}^S = \{ \hat{\mathbf{b}}^S_i \}_{i=1}^{L^{S}}$ from the student model and $ \hat{B}^T = \{ \hat{\mathbf{b}}^T_j \}_{j=1}^{L^{T}}$ from the teacher model, we extract their 3D RoI features $\hat{F}^S = \{ \hat{\mathbf{f}}^S_i \}_{i=1}^{L^{S}}$ and $ \hat{F}^T = \{ \hat{\mathbf{f}}^T_j \}_{j=1}^{L^{T}}$ from the backbones of their detectors. For each $\hat{\mathbf{b}}^S_i$, we follow Eq.~\ref{eq:findmaxiou} to find $ \hat{\mathbf{b}}^T_m$ in $\hat{B}^T$ that has the largest IoU with $\hat{\mathbf{b}}^S_i$.  If \( \text{IoU}(\hat{\mathbf{b}}_i^S, \hat{\mathbf{b}}_m^T) \geq \eta \), we obtain a matched pair $(\hat{\mathbf{b}}_i^S,\hat{\mathbf{b}}_m^T)$. Finally, we obtain a set of matched pairs $\hat{B}^{match} = \{ (\hat{\mathbf{b}}^S_i,\hat{\mathbf{b}}^T_i) \}_{i=1}^G$ of $\hat{B}^{S}$, where $G\leq L^{S}$. For each pair in $\hat{B}^{match}$, we align their 3D features via:

\begin{equation}
    \mathcal{L}_\text{ST} = \frac{1}{G} \sum_{i=1}^{G} (1 - \text{sim}(\hat{\mathbf{f}}^S_i , \hat{\mathbf{f}}^T_i))
\end{equation}
where $\text{sim}(\cdot,\cdot)$ is the cosine similarity. 

The loss for training the student model in the self-training stage is a combination of the detector's original loss $\mathcal{L}_{\text{det}}$, $\mathcal{L}_{\text{align}}^{\text{img}}$ in Eq.~\ref{eq:alignimage}, $\mathcal{L}_{\text{align}}^{\text{text}}$ in Eq.~\ref{eq:aligntext}, and the above $\mathcal{L}_{\text{ST}}$:
\begin{equation}
\mathcal{L_{\text{student}}} = \mathcal{L}_\text{det} + \alpha \mathcal{L}_\text{align}^{\text{text}} + \beta \mathcal{L}_\text{align}^{\text{img}} + \gamma \mathcal{L_{\text{ST}}}
\label{eq:loss-self-training}
\end{equation}

\subsection{Enhancing 3D Pseudo Labels with 3D Boxes from Images}
\label{sec:Enhancing3DPseudoLabelswith3DBoxesfromImages}
To generate 3D pseudo labels from images, we follow~\cite{zhang2024approaching} and obtain a set of 3D bounding boxes $\hat{B}^{3D}_{img} = \{\hat{\mathbf{b}}^{3D}_i\}_{i=1}^{H}$ based on 2D detection bounding boxes from the off-the-shelf open-vocabulary 2D detector GroundingDINO~\cite{liu2024grounding}. During the self-training stage, we select 3D pseudo labels from $\hat{B}^{3D}_{img}$ based on their distances and overlaps with pseudo labels generated by the teacher model. The selection can be formulated as:
\begin{equation}
\begin{aligned}
    \hat{B}_{\text{img}}^{\text{sel}} = \{ \hat{\mathbf{b}}_i^{3D} \mid  & \hat{\mathbf{b}}_i^{3D} \in \hat{B}_{img}^{3D}, \  dist(\hat{\mathbf{b}}_i^{3D}) \geq \tau, \\
    & \max_{\hat{\mathbf{b}}_j \in B_{teacher}} \left( \text{IoU}(\hat{\mathbf{b}}_i^{3D}, \hat{\mathbf{b}}_j) \right) \leq \xi
    \Big\}.
\end{aligned}
\end{equation}
where $B_{teacher}$ contains the pseudo labels generated by the teacher model, $dist(\cdot)$ obtains the distance of a 3D bounding box. The final set of pseudo labels is $B_{\text{pseudo}} = \hat{B}_{\text{img}}^{\text{sel}} \cup B_{\text{teacher}}$.

\section{Experiment}
\subsection{Experimental Settings}
\begin{table*}[ht]
\renewcommand{\arraystretch}{1.2}
    \resizebox{\textwidth}{!}{
        \begin{tabular}{c|l|c|c|c|c|c|c}
            \hline
            \multirow{2}{*}{Task} & \multirow{2}{*}{Method} & \multicolumn{2}{c|}{PV-RCNN~\cite{shi2020pv}} & \multicolumn{2}{c|}{PointPillars~\cite{lang2019pointpillars}} & \multicolumn{2}{c}{SECOND-IoU~\cite{yang2021st3d}} \\
            \cline{3-8}
            & & $\text{AP}_\text{{BEV}}$/$\text{AP}_\text{{3D}}$ & Closed Gap & $\text{AP}_\text{{BEV}}$/$\text{AP}_\text{{3D}}$ & Closed Gap & $\text{AP}_\text{{BEV}}$/$\text{AP}_\text{{3D}}$ & Closed Gap \\
            \hline
            \multirow{8}{*}{N$\rightarrow$K} 
            & Source Only & 68.2/37.2 & -/- & 22.8/0.5 & -/-   & 51.8/17.9 & -/-\\
            & SN$^\dagger$~\cite{wang2020train} & 60.5/49.5 & +36.8\%/+27.1\% & 39.3/2.0 & +26.6\%/+2.1\%  & 59.7/37.6 & +25.1\%/+35.4\% \\
            \cline{2-8}
            & ST3D~\cite{yang2021st3d} & 78.4/70.9 & +49.0\%/+74.3\% & 60.4/11.1 & +60.6\%/+14.9\% & 75.9/54.1 & +76.6\%/+59.5\% \\
            & DTS~\cite{hu2023density} & 83.9/71.8 & +75.8\%/+76.4\% & 79.5/51.8 & +91.5\%/+72.2\% & 81.4/66.6 & +94.0\%/+87.6\% \\
            & CMDA~\cite{chang2024cmda} & 84.9/75.0 & +80.3\%/+83.4\% & -/- & -/- & 82.1/69.0 & +96.2\%/+92.0\% \\
            & CMT~\cite{chen2024cmt} & 85.8/75.5 & +84.6\%/+84.5\% & 81.4/57.6 & +94.5\%/+80.3\% & 83.0/68.1 & +99.0\%/+90.3\% \\
            & GroupEXP-DA~\cite{shin2024towards} & -/- & -/- & \textbf{81.9}/52.8 & \textbf{+95.3\%}/+73.6\% & 81.5/68.2 & +94.3\%/+90.5\% \\
            & Ours & \textbf{86.8}/\textbf{78.1} & \textbf{+89.4\%/+90.3\%} & \textbf{81.9/60.4} & \textbf{+95.3\%/+84.2\%} & \textbf{84.8/69.8} & \textbf{+104.8\%/+93.3\%} \\
            \cline{2-8}
            & Oracle & 89.0/82.5 & -/- & 84.8/71.6 & -/- & 83.3/73.5 & -/- \\
            \hline
            \multirow{5}{*}{W$\rightarrow$K} 
            & Source Only & 61.2/22.0 & -/- & 47.8/11.5 & -/-  & 67.6/27.5 & -/- \\
            & SN$^\dagger$~\cite{wang2020train} & 79.8/63.6 & +66.9\%/+68.7\% & 27.4/6.4 & -55.1\%/-8.5\% & 79.0/59.2 & +72.3\%/+69.0\% \\
            \cline{2-8}
            & ST3D~\cite{yang2021st3d} & 84.1/64.8 & +82.4\%/+70.7\% & 58.1/23.2 & +27.8\%/+19.5\% & 82.2/61.8 & +93.0\%/+74.7\% \\
            & DTS~\cite{hu2023density} & 86.4/68.1 & +90.6\%/+76.2\% & 76.1/50.2 & +76.5\%/+64.4\% & 85.8/71.5 & +115.9\%/+95.7\% \\
            & CMT~\cite{chen2024cmt} & 85.9/\textbf{74.5} & +88.8\%/\textbf{+86.8\%} & 77.4/51.5 & +80.0\%/+66.6\% & 85.2/72.1 & +112.1\%/+97.0\% \\
            & GroupEXP-DA~\cite{shin2024towards} & -/- & -/- & 78.4/54.1 & +82.7\%/+70.9\% & \textbf{86.9/73.7} & \textbf{+122.9\%/+100.4\%} \\
            & Ours & \textbf{87.6}/72.7 & \textbf{+95.0\%}/+83.8\% & \textbf{81.4/56.8} & \textbf{+90.8\%/+75.4\%} & \textbf{86.9}/70.4 & \textbf{+122.9\%}/+93.3\% \\
            \cline{2-8}
            & Oracle & 89.0/82.5 & -/- & 84.8/71.6 & -/- & 83.3/73.5 & -/- \\
            \hline
            \multirow{9}{*}{W$\rightarrow$N} 
            & Source Only & 34.5/21.5 & -/- & 27.8/12.1 & -/-  & 32.9/17.2 & -/- \\
            & SN$^\dagger$~\cite{wang2020train} & 34.2/22.3 & -1.5\%/+4.8\% & 28.3/13.0 & +2.4\%/+4.7\% & 33.2/18.6 & +1.7\%/+7.5\% \\
            \cline{2-8}
            & ST3D~\cite{yang2021st3d} & 36.4/23.0 & +10.3\%/+8.8\% & 30.6/15.6 & +13.2\%/+18.2\% & 35.9/20.2 & +15.9\%/+16.7\% \\
            & L. D.~\cite{wei2022lidar} & 43.3/25.6 & +47.3\%/+24.0\% & 40.2/19.1 & +58.4\%/+36.5\% & 40.7/22.9 & +41.1\%/+32.2\% \\
            & DTS~\cite{hu2023density} & 44.0/26.2 & +51.1\%/+27.5\% & 42.2/21.5 & +67.9\%/+49.0\% & 41.2/23.0 & +43.7\%/+32.8\% \\
            & CMDA~\cite{chang2024cmda} & 44.4/26.4 & +53.2\%/+28.7\% & -/- & -/- & 42.8/\textbf{24.6} & +52.1\%/\textbf{+41.8\%} \\
            & CMT~\cite{chen2024cmt} & 41.7/26.4 & +38.7\%/+28.7\% & 38.0/21.1 & +48.1\%/+46.9\% & 40.8/24.3 & +41.6\%/+40.1\% \\
            & GroupEXP-DA~\cite{shin2024towards} & -/- & -/- & \textbf{44.3/22.2} & \textbf{+77.8\%/+52.6\%} & 43.8/24.4 & +57.4\%/+40.7\% \\
            & Ours & \textbf{45.5/27.0} & \textbf{+59.1\%/+32.2\%} & 42.5/21.5 & +69.3\%/+49.0\% & \textbf{45.1/24.6} & \textbf{+64.2\%/+41.8\%} \\
            \cline{2-8}
            & Oracle  & 53.1/38.6 & -/- & 49.0/31.3 & -/- & 51.9/34.9 & -/- \\
            \hline
        \end{tabular}
    }
    \centering
    \caption{Comparison with state-of-the-art approaches. ``Source Only" means the pre-trained model in the source domain is directly applied to the target domain. ``Oracle" indicates that the detector is trained with the labeled target domain data. $\dagger$ means SN is weakly supervised with target domain statistics. The best performances are in bold.}
    \label{tab:domain_adaptation}
\vspace{-7pt}
\end{table*}

\textbf{Datasets.} We evaluate our proposed MMAssist on three widely used 3D object detection datasets for autonomous driving: Waymo~\cite{sun2020scalability}, nuScenes~\cite{caesar2020nuscenes}, and KITTI~\cite{geiger2012we}. The point cloud data of Waymo dataset is captured by five LiDAR sensors, i.e., one 64-beam LiDAR and four 200-beam LiDARs, and the image data of Waymo is collected with five surrounding cameras. The nuScenes dataset provides point clouds captured by a 32-beam roof LiDAR, and its image data is collected by six surrounding cameras. KITTI uses a 64-beam LiDAR to collect point clouds, and it provides a pair of images captured by stereo cameras. Each dataset has its unique characteristics in sensor configuration, environment, etc. We consider each dataset as a separate domain and evaluate our MMAssist by adapting detectors with it across domains. We follow~\cite{hu2023density} and evaluate our MMAssist with the following domain adaptation tasks:  Waymo $\rightarrow$ nuScenes (W $\rightarrow$ N), Waymo $\rightarrow$ KITTI (W $\rightarrow$ K), and nuScenes $\rightarrow$ KITTI (N $\rightarrow$ K).

\noindent\textbf{Evaluation metric.}
Following~\cite{hu2023density}, we adopt the KITTI evaluation metric and perform evaluations on the commonly used car category (the vehicle category in Waymo). We report the average precision (AP) with respect to BEV IoUs and 3D IoUs over 40 recall positions with IoU threshold of 0.7.  The domain adaptation metric Closed Gap~\cite{yang2021st3d} is reported as well, and it is defined as $Closed\ Gap = \frac{\text{AP}_{model}\ -\ \text{AP}_{source}}{\text{AP}_{oracle}\ - \text{AP}_{source}}\times 100\%$. 

\noindent\textbf{Implementation details.}
\label{subsec:Experimental Settings}
The effectiveness of MMAssist is validated with SECOND-IoU~\cite{yang2021st3d}, PV-RCNN~\cite{shi2020pv}, and PointPillars~\cite{lang2019pointpillars}. We adopt the training settings of the popular point cloud detection codebase OpenPCDet~\cite{openpcdet2020} to pre-train the detectors in the source domain with feature alignment. The weights $\alpha$ and $\beta$ for the losses of feature alignment with image and text features are both set to 0.3 in the pre-training stage and both are set to 0.03 in the self-training stage.  The weight $\gamma$ of the feature alignment loss between teacher and student branches is set to 0.1. The IoU threshold 
 $\mu$ used to match the 3D predictions and labels is set to 0.5. For the self-training stage in the target domain, we use Adam~\cite{kingma2014adam} and one cycle scheduler to fine-tune the detectors for 30 epochs, and the initial learning rate is set to $1.5 \times 10^{-3}$. The smoothing coefficient  $\epsilon$ of EMA is set to 0.999. The IoU threshold $\eta$ for matching the predictions of the teacher and student models is set to 0.5. The IoU threshold $\xi$ and distance threshold $\tau$ used for selecting 3D labels generated from images are set to 0.5 and 30, respectively. 
 
 \noindent\textbf{Speed comparison.} Since the feature alignment is only performed in the training stage, our approach adds little extra computation burden to the test stage.  With a single RTX 4090 GPU, the test speeds of the baseline SECOND-IoU, PV-RCNN, PointPillars, and their MMAssist-enhanced versions (in the parentheses) are 52.04 FPS (51.40 FPS), 6.67 FPS (6.65 FPS), and 82.55 FPS (79.43 FPS), respectively, where FPS is short for frames per second.

\subsection{Comparison with State-of-the-Art}
\label{sec:main-results}
We compare our approach MMAssist with SN~\cite{wang2020train}, ST3D~\cite{yang2021st3d}, L.D~\cite{wei2022lidar},  DTS~\cite{hu2023density}, CMDA~\cite{chang2024cmda}, CMT~\cite{chen2024cmt} and GroupEXP-DA~\cite{shin2024towards}. As shown in Table~\ref{tab:domain_adaptation}, when combined with PV-RCNN,  on N $\rightarrow$ K, MMAssist surpasses the second-best method CMT by $1.0\%$ $\text{AP}_{\text{BEV}}$ and $2.6\%$ $\text{AP}_{\text{3D}}$. On W $\rightarrow$ K, MMAssist is $1.7\%$ $\text{AP}_{\text{BEV}}$ better than CMT while CMT is $1.8\%$ $\text{AP}_{\text{3D}}$ better than ours. On W $\rightarrow$ N, our approach outperforms the second-best CMDA by $1.1\%$ $\text{AP}_{\text{BEV}}$ and $0.6\%$ $\text{AP}_{\text{3D}}$. The performance of MMAssist with PointPillars is also remarkable. On N $\rightarrow$ K, MMAssist and GroupEXP-DA both achieve the best $81.9\%$ $\text{AP}_{\text{BEV}}$, but our approach is $7.6\%$ $\text{AP}_{\text{3D}}$ better than  GroupEXP-DA. Ours also outperforms second-best method GroupEXP-DA by $3.0\%$ $\text{AP}_{\text{BEV}}$ and $2.7\%$ $\text{AP}_{\text{3D}}$ on W $\rightarrow$ K. On W $\rightarrow$ N,  GroupEXP-DA performs best and excels our approach by $1.8\%$ $\text{AP}_{\text{BEV}}$ and $0.7\%$  $\text{AP}_{\text{3D}}$. In terms of SECOND-IoU,  on N $\rightarrow$ K, our approach outperforms the overall second best CMT by $1.8\%$ $\text{AP}_{\text{BEV}}$ and $1.7\%$ $\text{AP}_{\text{3D}}$. On W $\rightarrow$ K, both MMAssist and GroupEXP-DA show the best performance regarding $\text{AP}_{\text{BEV}}$, but GroupEXP-DA outperforms our approach by $3.3\%$ $\text{AP}_{\text{3D}}$. On W $\rightarrow$ N, our approach surpasses the second-best  $\text{AP}_{\text{BEV}}$ of  GroupEXP-DA by $1.3\%$ and shares the best $\text{AP}_{\text{3D}}$ with CMDA.

\subsection{Ablation Studies}
In this section, we conduct ablation studies on the Waymo $\rightarrow$ KITTI task with PointPillars~\cite{lang2019pointpillars}.

\begin{table}[htbp]
\centering
\renewcommand{\arraystretch}{1.2}
\setlength{\tabcolsep}{2pt}
\scalebox{1}{
\begin{tabular}{c|cc|cccc|cc}
\hline
  \multirow{2}{*}{Method}& \multicolumn{2}{c|}{Pre-Training} & \multicolumn{4}{c|}{Self-Training} & \multirow{2}{*}{$\text{AP}_\text{{BEV}}$} & \multirow{2}{*}{$\text{AP}_\text{{3D}}$} \\ \cline{2-7}
      & IA    &TA     & CAM             & IA       &TA      &    STA &                     &                        \\ \hline
(a)   &   &   &   &   &  & & 76.7  & 52.7           \\
(b)   &   & &\checkmark & & & & 79.1 & 53.2        \\
(c)   &\checkmark & &\checkmark &\checkmark & & & 80.5 & 54.4                                            \\
(d)   & &\checkmark &\checkmark & &\checkmark & & 80.0              & 54.5                                    \\
(e)   &\checkmark &\checkmark &\checkmark &\checkmark &\checkmark & & 80.9     & 55.7               \\
(f)   &\checkmark &\checkmark &\checkmark &\checkmark &\checkmark &\checkmark & 81.4  & 56.8           \\
\hline
\end{tabular}
}
\caption{Validation of the effectiveness of each component of MMAssist. Please refer to the text for the meanings of ``IA", ``TA", ``CAM", and ``STA".}
\label{table:ablations}
\vspace{-5pt}
\end{table}
\noindent\textbf{Main ablation.}
We analyze the impact of each component of our MMAssist with the results in Table~\ref{table:ablations}. ``IA" represents aligning 3D features with image features, ``TA" represents aligning 3D features with text features,  ``CAM" means incorporating pseudo labels generated from 2D images, and ``STA" indicates aligning the 3D features between the student and teacher branches. (a) is the baseline, which is the basic student-teacher approach with only beam re-sampling~\cite{hu2023density}.  (b) incorporates new pseudo labels generated from images during self-training in the target domain. It can be observed that the newly added pseudo labels improve both $\text{AP}_\text{{3D}}$ and $\text{AP}_\text{{BEV}}$. (c) and (d) further align the 3D features with either image features or text features. The results show that compared with (b), both ``IA" and ``TA" bring improvements in $\text{AP}_\text{{BEV}}$ and $\text{AP}_\text{{3D}}$. (e) aligns the 3D features with both image and text features, and the performance is better than both (c) and (d). (f) is our full MMAssist, and we can see that compared with (e), ``STA" can further improve the performance. These experiments effectively demonstrate the usefulness of each component. It is worth noting that compared with the vanilla PointPillars, which is (a), we add a second stage to make use of the aligned features, but simply adding a second stage does not bring about notable performance gain, the $\text{AP}_\text{{BEV}}$ increases to $78.3\%$ while $\text{AP}_\text{{3D}}$ decreases to $51.9\%$.

\begin{table}[htbp]
    \centering
\renewcommand{\arraystretch}{1.2}
        \begin{tabular}{c|c|cc}
            \hline
            Pre-Training & Self-Training & \multicolumn{1}{l}{$\text{AP}_\text{{BEV}}$} & \multicolumn{1}{l}{$\text{AP}_\text{{3D}}$} \\
            \hline
             &  & 79.1 & 53.2  \\
             IA+TA&  & 79.7 & 55.3 \\
             & IA+TA & 79.8 & 54.6 \\
            IA+TA & IA+TA & 81.4 & 56.8 \\
            \hline
        \end{tabular}
    \caption{Analysis of the bridge effect of image and  text features.  ``IA" and ``TA" have the same meaning as Table~\ref{table:ablations}.}
    \label{tab:align_text}
\end{table}

\noindent\textbf{Bridge effect of image and text features for 3D feature alignment.} 
To further verify that the usefulness of image and text features mainly comes from them serving as bridges between the 3D features of the source domain and the target domain, on top of (b) in Table~\ref{table:ablations}, we add ``IA" and ``TA" only to the pre-training stage and only to the self-training stage, respectively. The results in Table~\ref{tab:align_text} show that when ``IA" and ``TA" are added to both the pre-training and the self-training stages, the performance gain is notably larger than only adding them to one of the two training stages.

\begin{table}[htbp]
    \centering 

    \begin{minipage}[t]{0.18\textwidth} 
        \centering
        \renewcommand{\arraystretch}{1.2}
        \setlength{\tabcolsep}{2pt} 
        \scalebox{1}{
        \begin{tabular}{c|c}
            \hline
            Method & $\text{AP}_\text{{BEV}}$/$\text{AP}_\text{{3D}}$ \\ \hline
            Concat     & 80.5/55.4  \\
            Sum       & 78.8/50.8  \\ 
            WSum  & 81.4/56.8    \\ \hline
        \end{tabular}}
        \captionof{table}{Comparison of different feature fusion methods.} 
        \label{tab:fusion}
    \end{minipage}
    \hfill 
    \begin{minipage}[t]{0.27\textwidth} 
        \centering
        \renewcommand{\arraystretch}{1.2}
        \setlength{\tabcolsep}{2pt} 
        \scalebox{1}{
        \begin{tabular}{c|c}
            \hline
            Method & $\text{AP}_\text{{BEV}}$/$\text{AP}_\text{{3D}}$ \\ \hline
            Qwen2-VL + SLIP    & 80.2/55.7  \\ 
            LLaVA + LLaMA     & 80.6/52.7  \\
            LLaVA + SLIP      & 81.4/56.8    \\ \hline
        \end{tabular}}
        \captionof{table}{Comparison of text feature generation methods.} 
        \label{tab:lvlm}
    \end{minipage}

\end{table} 
\noindent\textbf{Comparison of different feature fusion methods.} 
To validate the effectiveness of our weighted element-wise sum approach (WSum) to fuse 3D features with aligned features, we compare our approach with element-wise sum (Sum) and concatenation (Concat) in Table~\ref{tab:fusion}. ``Concat" concatenates the 3D features, image-aligned feature, and text-aligned feature and processes it with MLP. ``Sum" directly sums the 3D features and the aligned features after changing their dimensionalities. The result shows that our approach is more effective.

\noindent\textbf{Comparison of different text feature generation methods.} For choice of pre-trained LVLM, we test LLaVA (v1.5-13B)~\cite{liu2023visual} and Qwen2-VL (Qwen2-VL-2B)~\cite{wang2024qwen2}, the results in Table~\ref{tab:lvlm} show that LLaVA is a better choice for MMAssist. Regarding the choice of the pre-trained text encoder, we test SLIP~\cite{mu2022slip} and LLaMA (Llama-2-7B)~\cite{touvron2023llama}. The results reveal that SLIP is a better choice for our approach.

\begin{table}[t!]
\small
\centering
\renewcommand{\arraystretch}{1.2}
\begin{tabular}{c|ccc}
\hline
\multirow{2}{*}{Method} & \multicolumn{3}{c}{$\text{AP}_\text{BEV}$ / $\text{AP}_\text{3D}$} \\ 
\cline{2-4}
 & 0-30m & 30-60m & 60-150m \\ 
\hline
Baseline       & 81.0 / \textbf{58.5}  & 50.7 / 24.6  & 3.2 / 0.2  \\ 
Baseline+CAM   & \textbf{81.4} / 58.4  & \textbf{51.9} / \textbf{27.9}  & \textbf{5.0} / \textbf{0.7}  \\ 
\hline
\end{tabular}
\caption{Effect of adding pseudo labels generated from images. ``CAM" has the same meaning as Table~\ref{table:ablations}.}
\label{table:distance}
\vspace{-5pt}
\end{table}

\noindent\textbf{Effect of incorporating 3D pseudo labels generated from images at different ranges.} To demonstrate the effectiveness of combining the pseudo labels generated from images and the pseudo labels generated by the teacher model, we evaluate ``Baseline"  and ``Baseline+CAM" at different distance ranges. ``Baseline" is (a) in Table~\ref{table:ablations}, and ``Baseline+CAM" is (b) in Table~\ref{table:ablations}, which uses the pseudo labels generated from images during the self-training. The results in Table~\ref{table:distance} show pseudo labels generated from images just have a very small impact in the range of 0-30m, but they improve the performance in the range of 30-60m by $3.3\%$ $\text{AP}_{3D}$ and $1.2\%$ $\text{AP}_{\text{BEV}}$. The performance in the range of 60-150m can also be improved by $0.5\%$ $\text{AP}_{3D}$ and $1.8\%$ $\text{AP}_{\text{BEV}}$. This is consistent with our pseudo label combination strategy.

\section{Conclusion and Limitation}
\label{sec:conclusion}
In this paper, we propose using multi-modal information to assist in the task of unsupervised domain adaptation on point cloud 3D object detection (3D UDA). Image and text features are used as bridges to align the 3D features between the source and target domains. The pseudo labels for self-training are enhanced with pseudo labels estimated from 2D detection results as well. Comprehensive evaluations demonstrate the effectiveness of our approach. A limitation of our work is that features are aligned at the instance level that is short of semantic information.  We will investigate the usefulness of global semantic information for assisting in 3D UDA in our future work.       
\section{Acknowledgment}
\label{sec:ack}
This work is in part supported by Natural Science
Foundation of Henan Province (No. 252300421503 and 242300420270). 

\bibliography{aaai2026}

\clearpage

\section{Supplementary Material}
\subsection{Extracting 3D Features from  3D Detectors}
\label{sec:3D-feature-extraction}
We integrate our approach into three 3D object detectors SECOND-IoU~\cite{yang2021st3d}, PV-RCNN~\cite{shi2020pv}, and PointPillars~\cite{lang2019pointpillars}. In the following, we describe the process of extracting the 3D features of 3D bounding boxes for each detector. 

\noindent\textbf{SECOND-IoU.}
 SECOND-IoU~\cite{yang2021st3d} predicts the classification and 3D bounding box information of objects from the BEV feature maps before estimating the IoU score. For each predicted 3D box $\mathbf{b}=(x,y,z,l,w,h,\theta)$, we project it to the BEV feature maps and use RoI pooling and max pooling to extract the 3D feature, which is used for feature alignment.
 
\noindent\textbf{PV-RCNN.}
PV-RCNN~\cite{shi2020pv} generates 3D proposals based on the BEV feature maps in the first stage. In the second stage, for each 3D proposal $\mathbf{b}=(x,y,z,l,w,h,\theta)$, it extracts the RoI feature of $\mathbf{b}$ by aggregating keypoint features for further processing. We convert the RoI features of 3D proposals to 3D features for feature alignment with max pooling.  

\noindent\textbf{PointPillars.}
PointPillars~\cite{lang2019pointpillars} directly predicts the final 3D bounding boxes by performing object classification and bounding box regression on the BEV feature maps. We treat the output of the original PointPillars as the initially detected 3D bounding boxes. For each initial 3D bounding box $\mathbf{b}=(x,y,z,l,w,h,\theta)$, we project it to the BEV feature maps and extract 3D features for feature alignment with RoI pooling and max pooling.

\subsection{More Comparisons of Point Cloud, Image, and Text}
We present six more comparisons of point cloud, image, and object's text description generated by LLaVA~\cite{liu2023visual} of two objects having similar appearances in the images. (a) is from Waymo~\cite{sun2020scalability}, and (b) is from nuScenes~\cite{caesar2020nuscenes}. The figures are best viewed in color.

\begin{figure}[h]
    \centering
    \includegraphics[width=\columnwidth]{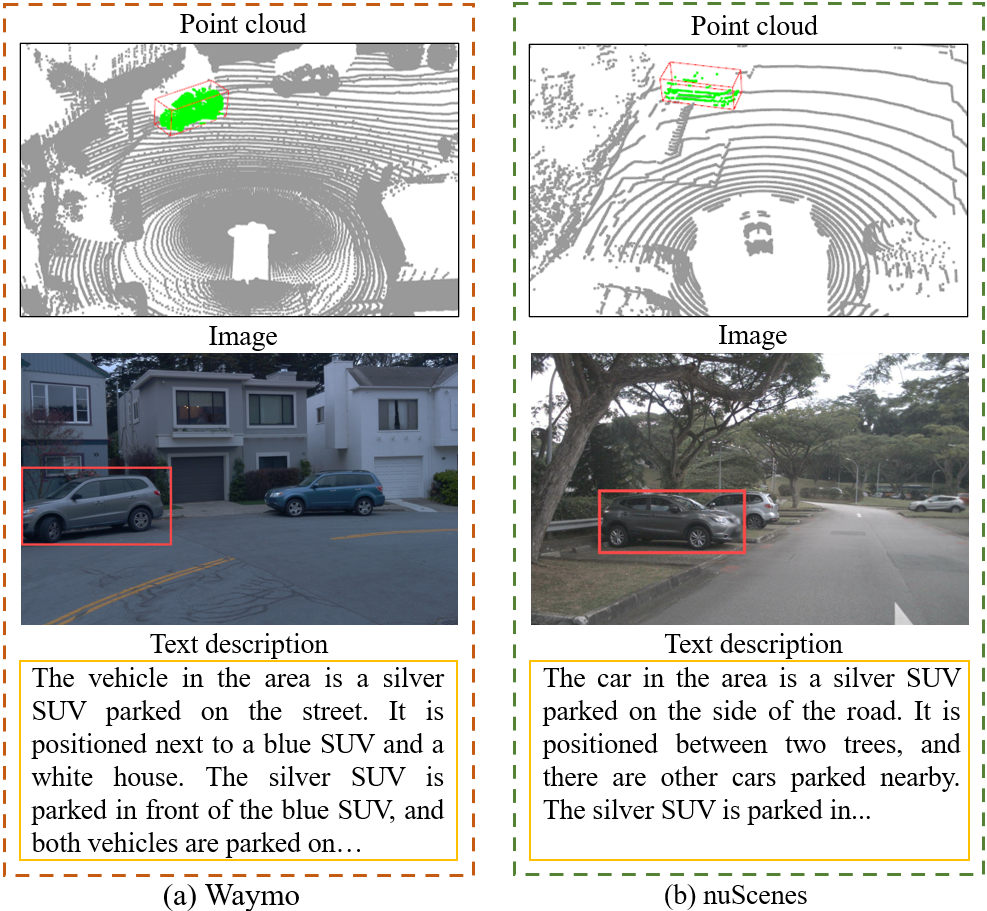}
    \vspace{-16pt}
    \caption{}
    \label{fig:fig2}
\end{figure}

\begin{figure}[h]
    \centering
    \includegraphics[width=\columnwidth]{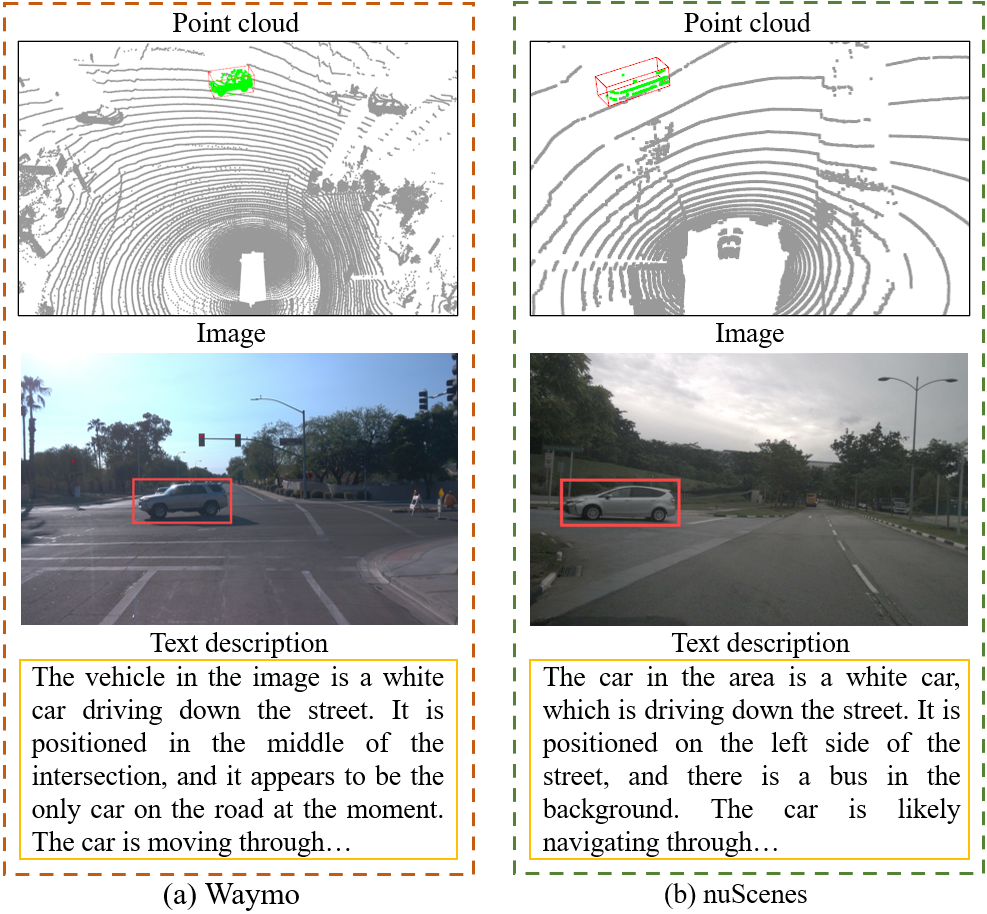}
    \vspace{-16pt}
    \caption{}
    \label{fig:fig3}
\end{figure}

\begin{figure}[h]
    \centering
    \includegraphics[width=\columnwidth]{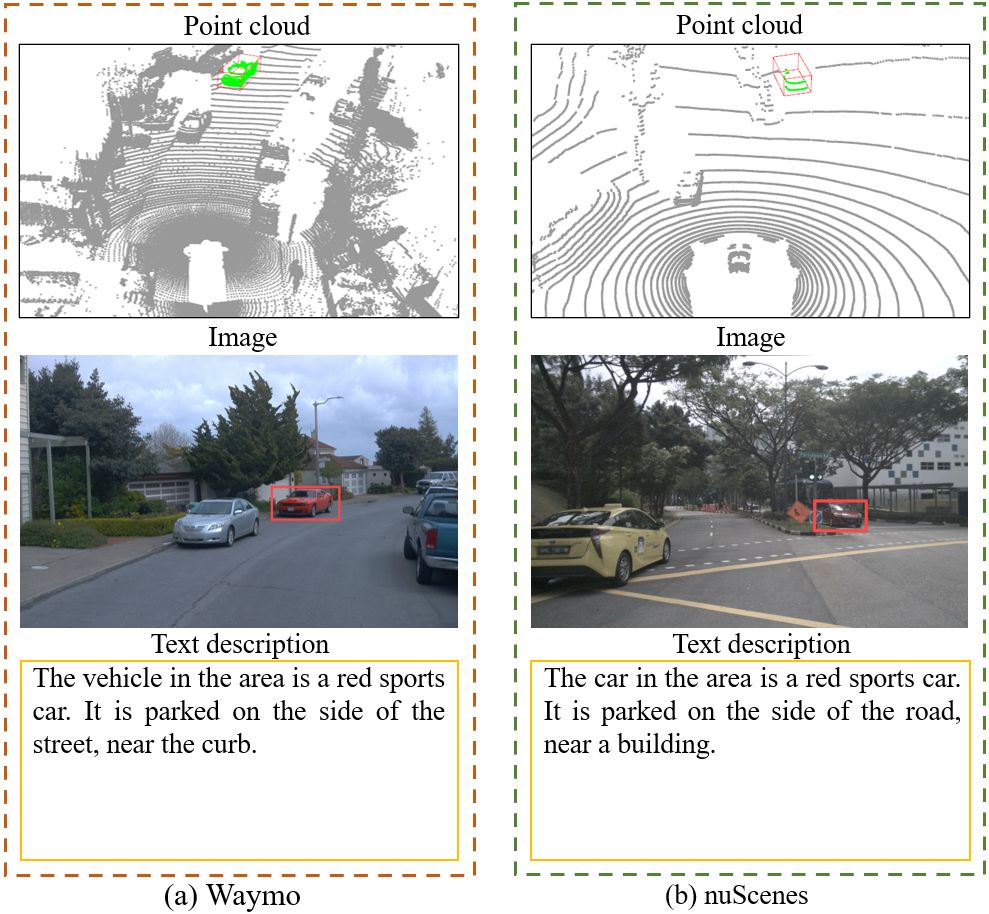}
    \vspace{-16pt}
    \caption{}
    \label{fig:fig4}
\end{figure}

\begin{figure}[h]
    \centering
    \includegraphics[width=\columnwidth]{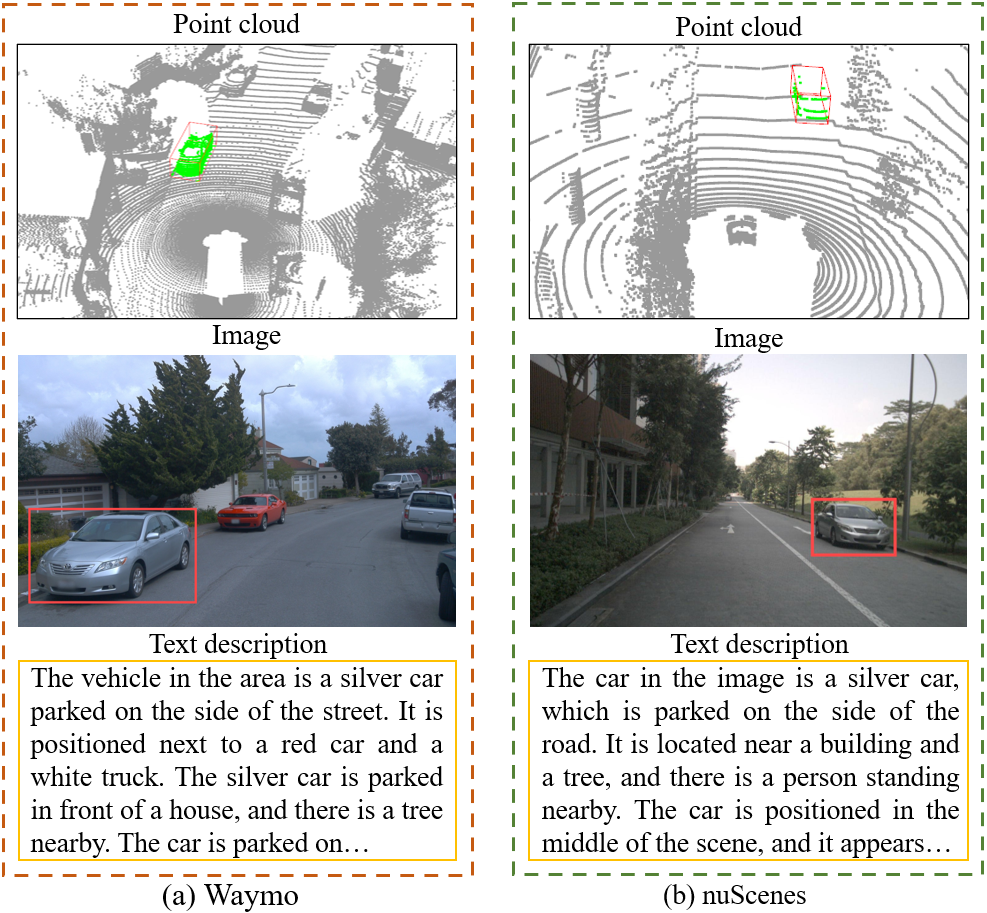}
    \vspace{-16pt}
    \caption{}
    \label{fig:fig5}
\end{figure}

\begin{figure}[h]
    \centering
    \includegraphics[width=\columnwidth]{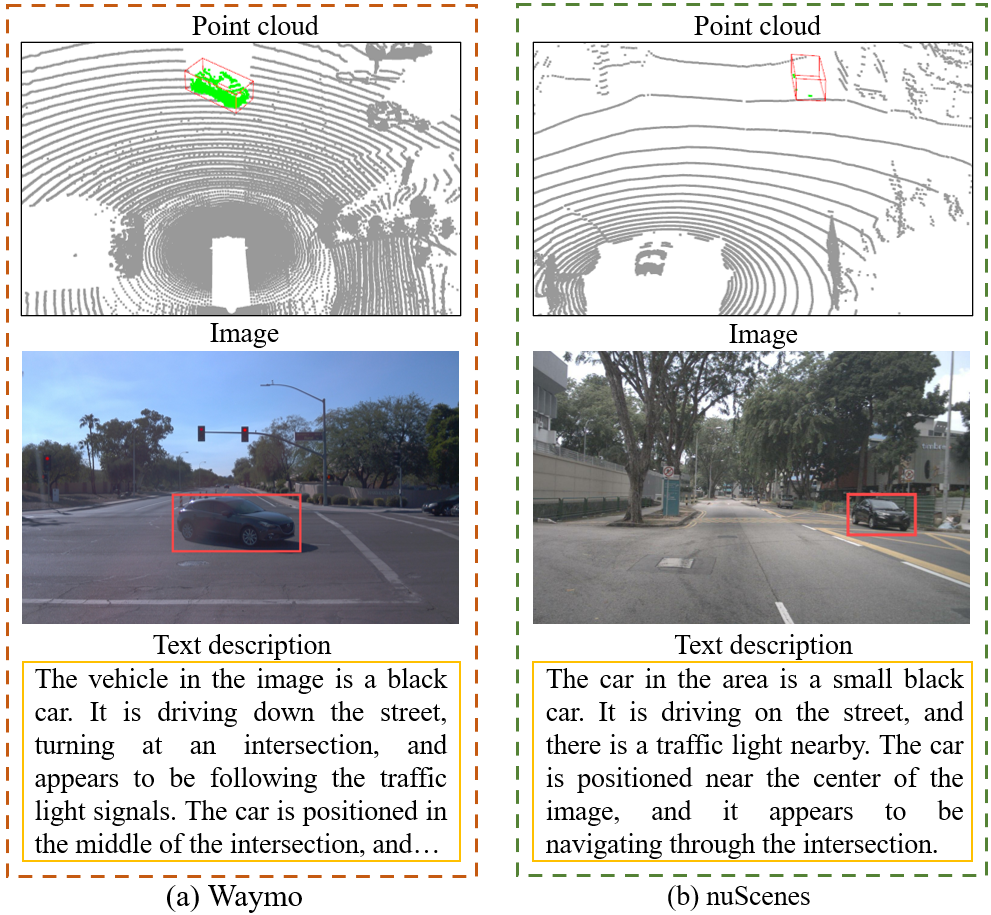}
    \vspace{-16pt}
    \caption{}
    \label{fig:fig1}
\end{figure}

\begin{figure}[h]
    \centering
    \includegraphics[width=\columnwidth]{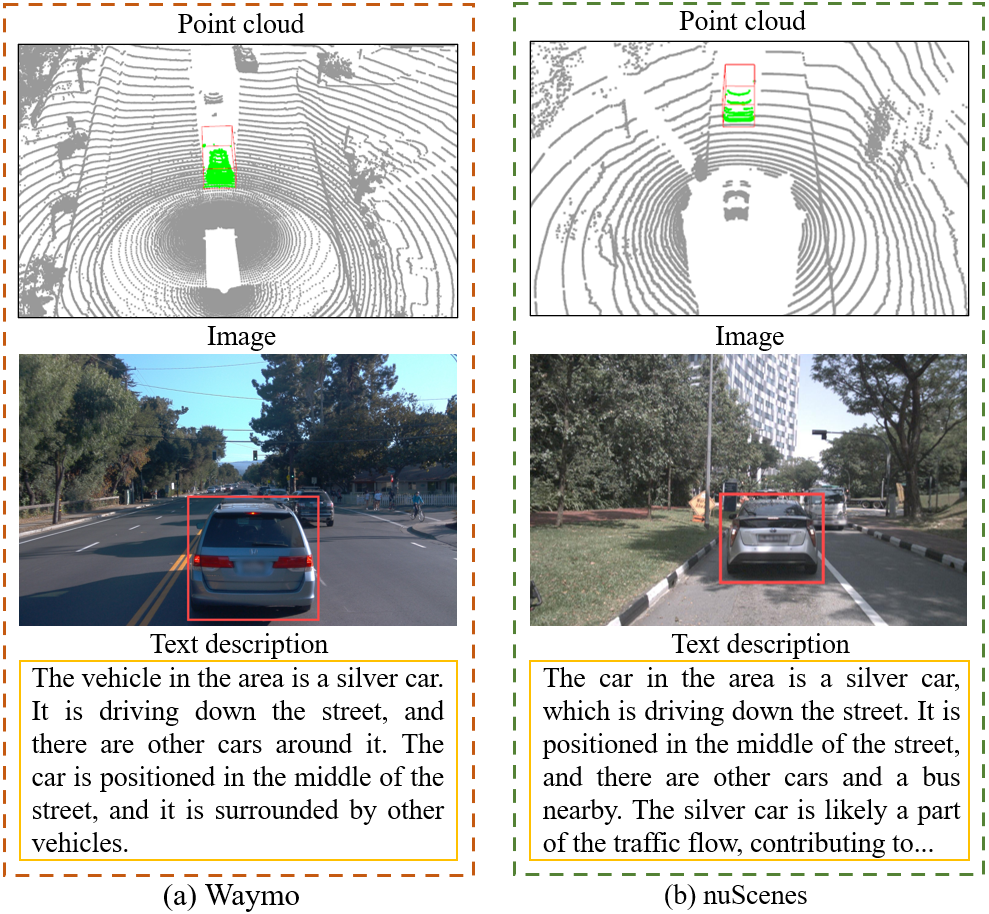}
    \vspace{-16pt}
    \caption{}
    \label{fig:fig6}
\end{figure}


\end{document}